\documentclass[lettersize,journal]{IEEEtran}
\usepackage{amsmath,amsfonts}
\usepackage{algorithmic}
\usepackage{algorithm}
\usepackage{array}
\usepackage[caption=false,font=normalsize,labelfont=sf,textfont=sf]{subfig}
\usepackage{textcomp}
\usepackage{stfloats}
\usepackage{url}
\usepackage{multirow}
\usepackage{booktabs}
\usepackage{indentfirst} 
\usepackage{verbatim}
\usepackage{enumitem}
\usepackage{graphicx}
\usepackage{cite}
\usepackage{hyperref}
\usepackage{cleveref}
\usepackage[normalem]{ulem}
\usepackage{xspace}
\usepackage[T1]{fontenc}
\useunder{\uline}{\ul}{} 
\begin{document}

\title{Style-Preserving Lip Sync via Audio-Aware Style Reference} 

\author{
        Weizhi~Zhong, 
        Jichang~Li, 
        Yinqi~Cai, 
        Ming~Li, 
        Feng~Gao,
        Liang~Lin, \IEEEmembership{Fellow,~IEEE,} 
        and Guanbin~Li
\thanks{This work was supported in part by the National Natural Science Foundation of China (NO.~62322608). (Corresponding authors are Feng Gao and Guanbin Li).}
\thanks{W.~Zhong, Y.~Cai, L.~Lin and G.~Li are with the School of Computer Science and Engineering, Sun Yat-sen University, Guangzhou, China. 
E-mail: zhongwzh5@mail2.sysu.edu.cn, caiyq27@mail2.sysu.edu.cn, linliang@ieee.org, liguanbin@mail.sysu.edu.cn}
\thanks{M.~Li is with the Shandong Inspur Database Technology Co., Ltd. E-mail: liming2017@inspur.com}
\thanks{F.~Gao is with the Peking University. E-mail: gaof@pku.edu.cn}
\thanks{J.~Li is with the Peng Cheng Laboratory, Shenzhen 518000, China. 
E-mail: li.jichang@pcl.ac.cn}
}

\markboth{Journal of \LaTeX\ Class Files,~Vol.~14, No.~8, May~2025}%
{Shell \MakeLowercase{\textit{et al.}}: A Sample Article Using IEEEtran.cls for IEEE Journals}


\maketitle

\begin{abstract}
Audio-driven lip sync has recently drawn significant attention due to its widespread application in the multimedia domain. 
Individuals exhibit distinct lip shapes when speaking the same utterance, attributed to the unique speaking styles of individuals, posing a notable challenge for audio-driven lip sync. 
Earlier methods for such task often bypassed the modeling of personalized speaking styles, resulting in sub-optimal lip sync conforming to the general styles. Recent lip sync techniques attempt to guide the lip sync for arbitrary audio by aggregating information from a style reference video, yet they can not preserve the speaking styles well due to their inaccuracy in style aggregation. 
This work proposes an innovative audio-aware style reference scheme that effectively leverages the relationships between input audio and reference audio from style reference video to address the style-preserving audio-driven lip sync. 
Specifically, we first develop an advanced Transformer-based model adept at predicting lip motion corresponding to the input audio, augmented by the style information aggregated through cross-attention layers from style reference video. Afterwards, to better render the lip motion into realistic talking face video, we devise a conditional latent diffusion model, integrating lip motion through modulated convolutional layers and fusing reference facial images via spatial cross-attention layers. Extensive experiments validate the efficacy of the proposed approach in achieving precise lip sync, preserving speaking styles, and generating high-fidelity, realistic talking face videos.
\end{abstract}


\section{Introduction}

\IEEEPARstart{A}{udio}-driven lip sync, also referred to as talking face generation, aims to generate a talking video having lip motion synchronized with the input audio.
This technology has garnered significant attention in the multimedia domain in recent years, spanning applications such as video translation~\cite{kr2019towards}, visual dubbing~\cite{xie2021towards}, virtual avatars~\cite{zhou2022dialoguenerf}, and digital humans~\cite{thies2020neural}. 
Traditionally, achieving accurate lip sync has primarily involved training subject-specific models with data collected from specific individuals. However, these techniques are hampered by the significant costs associated with subject-specific training and data collection processes. Consequently, we focus on addressing the task of audio-driven lip sync through a subject-generic strategy~\cite{wang2024styletalk++,ma2023styletalk,prajwal2020lip,zhong2023identity,zhou2021pose,shen2023difftalk}, which, by training once on a large-scale dataset, can be generalized to any individual without the need for further training or fine-tuning.

\begin{figure}
    \centering
    \includegraphics[width=\columnwidth]{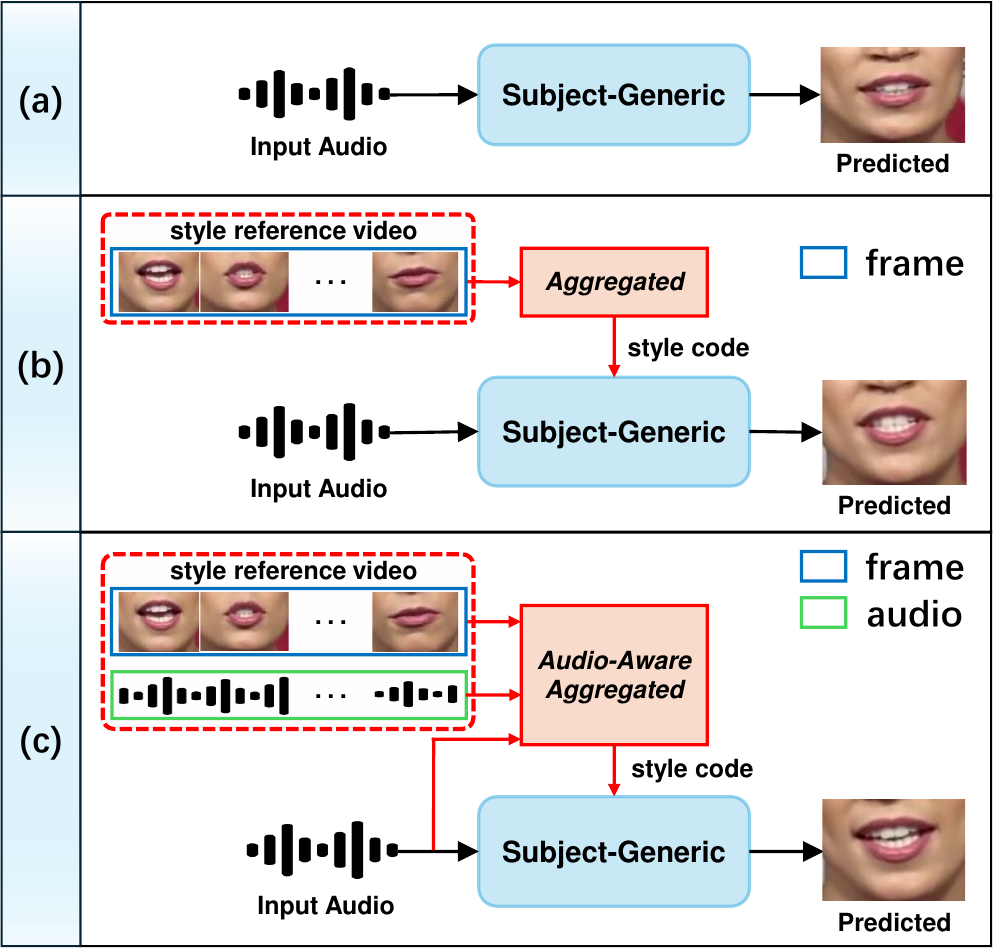}
    \vspace{-15pt}
    \caption{Conceptual comparisons between prior algorithms and the proposed \textbf{StyleSync}. 
    (a) Earlier methods overlooked incorporating reference information pertaining to individual speaking styles, giving rise to sub-optimal lip sync adhering to general style.
    (b) Recent algorithms aggregate a video-level style code from lip motion sequences of a style reference video to guide lip sync for arbitrary given audio, but such a style code proves inadequate for attaining accurate lip sync.
    (c) This work proposes a novel scheme of audio-aware style aggregation that capitalizes on relationships between input audio and the reference audio from style reference video, thereby enabling lip sync that preserves the styles of individuals better.
    }
    \label{fig:intro}
\end{figure}


In general, when individuals speak the same utterance, variations in their speaking styles lead to differences in lip shapes~\cite{fu2024mimic,ma2023styletalk,wang2024styletalk++,tan2024say,wu2021imitating,zhang2021flow,chen2023vast}, posing a challenge to accurate lip sync. 
However, most previous subject-generic approaches, such as~\cite{prajwal2020lip,zhong2023identity,zhou2021pose,shen2023difftalk,xie2021towards,wang2023progressive,kr2019towards}, have overlooked modeling the speaking style of individuals, as shown in Figure \ref{fig:intro}(a). 
These approaches merely predict lip motion from audio without incorporating any reference information regarding the speaking style of subject. As a result, these subject-generic approaches typically only achieve lip sync consistent with the general speaking styles learned from audio-visual datasets. There are still discrepancies between the lip sync conforming to general speaking styles and those customized to individual speaking styles.

To alleviate this issue, recent methods~\cite{ma2023styletalk,wang2024styletalk++,fu2024mimic,tan2024say,wu2021imitating,ma2023talkclip} have attempted to incorporate a short style reference video of the individual into the lip sync model as illustrated in Figure \ref{fig:intro}(b). 
They learn a video-level style code from the reference video to guide the lip sync for arbitrary audio. 
However, predicting lip shapes for different audio requires referencing different lip shapes from the style reference video. Therefore, the compact and static video-level style codes learned by these methods are still insufficient to achieve satisfactory lip sync.

Inspired by the observation from~\cite{li2024ae} that similar pronunciations produce similar lip shapes, this paper posits that within a style reference video, if a lip shape corresponds to the reference audio that is closer to the input audio, it should contribute more significantly to the style coding. Thus, we approach Audio-driven Lip Sync from a novel perspective and propose an innovative audio-aware style reference scheme. As shown in Figure \ref{fig:intro}(c), our method leverages the audio signals in the style reference video to aggregate distinct style codes for different input audios, facilitating Style-Preserving Lip Sync.

Specifically, we propose a novel Transformer-based architecture designed to predict lip motions from audio inputs, influenced by a style reference video. The process commences with the extraction of lip shapes from a style reference video utilizing the 3D Morphable Model (3DMM)~\cite{blanz1999morphable}. Subsequently, these lip shapes and their associated audio signals are encoded via dual Transformer encoders, yielding respective representations for reference lip and audio features.   
Input audio information is then integrated into a Transformer decoder to generate hidden features, which are further processed by a cross-attention layer within the decoder to aggregate style information from the style reference video. Herein, the reference audio and lip features serve as keys and values, respectively, while the hidden features derived from the input audio act as queries. Ultimately, the Transformer decoder outputs mouth-related expression parameters representing lip motion. 
To render realistic talking face videos from lip motion, this research devises a conditional latent diffusion model~\cite{rombach2022high} tailored for producing lip-synced faces. Specifically, the integration of mouth-related expression parameters into the latent diffusion model is achieved through modulated convolutional layers. This approach ensures that the synthesized lip shapes are 
aligned with those represented by mouth-related expression parameters. 
Additionally, the incorporation of reference facial image into the diffusion model is facilitated via spatial cross-attention layers, which enhance the model's capacity to capture subject-specific appearance details, thereby improving the fidelity of the generated talking faces.

The primary contributions of this study are summarized as follows.
\begin{itemize}
    \item We propose a novel audio-aware style reference scheme that leverages the intrinsic connections between input and reference audios to effectively aggregate speaking style information, thereby achieving style-preserving lip synchronization.
    \item We have devised an advanced Transformer-based lip sync model that integrates speaking style information from a style reference video to predict lip motions. Additionally, we introduced a novel conditional latent diffusion model to translate the predicted lip motions into lip-synced and realistic videos.
    \item Extensive experiments demonstrate that our proposed method achieves style-preserving lip synchronization, which effectively preserves the speaking style of the individual, resulting in realistic and high-fidelity talking face videos. 
\end{itemize}

\section{RELATED WORK}

\subsection{Audio-Driven Lip Sync}
With the increasing application in multimedia, audio-driven lip sync has attracted widespread attention in recent years. 
According to the training paradigm, methods for audio-driven lip sync can be categorized into subject-specific and subject-generic. 
Subject-specific methods\cite{li2024ae,guo2021ad,shen2023sd,du2023dae,yi2022predicting} need to collect data of the target subject for subject-specific training.
Although they can achieve lip sync conforming to the individual's speaking style, they can not be generalized to unseen subjects without further training, thus having limited application in the real world.   
In contrast, subject-generic approaches for lip sync, such as ~\cite{wang2024styletalk++,ma2023styletalk,prajwal2020lip,zhong2023identity,zhou2021pose,shen2023difftalk,wang2023progressive,wang2022anyonenet,ye2022audio}, have garnered more attention for their ability to infer lip sync for unseen subjects without further training. However, most subject-generic methods~\cite{prajwal2020lip,zhong2023identity,zhou2021pose,shen2023difftalk,xie2021towards,wang2023progressive,kr2019towards} neglect to model the speaking styles of individuals in the absence of reference information regarding individual speaking style.
For example, Wav2Lip~\cite{prajwal2020lip} pre-trains a lip synchronization discriminator on a large-scale audio-visual dataset to guide the training of a lip sync generator. 
Yet, it can only achieve lip sync matching the general speaking styles learned from the audio-visual dataset. 
Similarly, IP-LAP~\cite{zhong2023identity} proposes a two-stage talking face generation method based on landmark representation, using a transformer-based model to predict lip landmarks from input audio. However, due to the lack of reference information on individual speaking style, it can not generate lip motions that preserve the speaking style well. 
In this paper, we propose a method that further improves subject-generic methods, achieving more accurate lip sync that better preserves the individual's speaking style.

\subsection{Style-Guided Lip Sync}

Recent advancements in subject-generic methods, e.g.~\cite{ma2023styletalk,wang2024styletalk++,fu2024mimic,tan2024say,wu2021imitating,ma2023talkclip},  have made efforts to model the speaking styles of individuals through the integration of style reference videos. Among these, 
StyleAvatar~\cite{wu2021imitating} constructs a video-level style code from a style reference video based on the standard deviation of facial movements. 
Then, it proposes a latent-style-fusion model to synthesize stylized talking faces for arbitrary audio, imitating the talking style from the style code. 
Recently, StyleTalk~\cite{ma2023styletalk} introduces a framework for generating stylized one-shot talking heads. 
This framework encodes the expression parameters from all frames within a style reference video into style vectors. 
These vectors are then aggregated into a single video-level style code through a self-attention pooling layer. 
Specifically, this layer calculates the attention weight for each frame by utilizing a feed-forward network, subsequently aggregating the style vectors weighted by their respective attention weights into a video-level style code. 
The derived style code is then used to guide the lip sync for arbitrary input audio. 
Following StyleTalk~\cite{ma2023styletalk}, some later works~\cite{wang2024styletalk++,fu2024mimic,tan2024say,ma2023talkclip} adopt a similar style aggregation method to obtain a compact video-level style code.
Despite their efforts, a common limitation persists across these methods: the inability to well preserve the speaking style of the individual during the task of lip sync. 
The main reason for this challenge is that, to predict the lip shape for different input audio, the network needs to reference different lip shapes from the style reference video. In response to this challenge, our study proposes a novel audio-aware style reference scheme to preserve the speaking style better for improved lip sync.
\section{METHODOLOGY}

\begin{figure*}[h!]  
  \centering 
  \includegraphics[width=\linewidth]{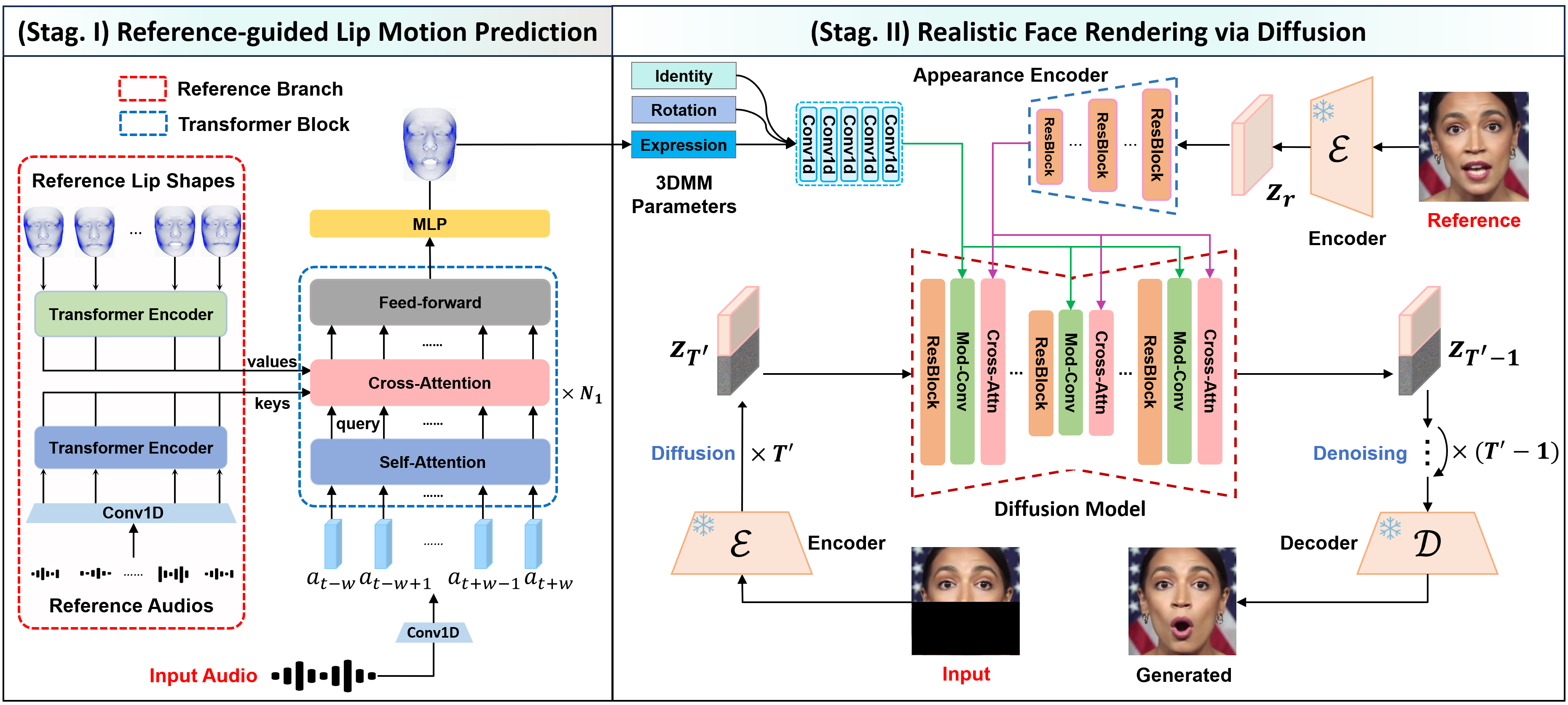} 
  \caption{An overview of the proposed \textbf{StyleSync}. 
StyleSync consists of two stages: Reference-guided Lip Motion Prediction and Realistic Face Rendering. \textbf{(1).} In Reference-guided Lip Motion Prediction, StyleSync utilizes a novel Transformer-based model to predict expression parameters related to lip motion. To facilitate explanation, we use the facial mesh to represent the mouth-related expression parameters. \textbf{(2).} In the stage of Realistic Face Rendering, StyleSync develops a conditional latent diffusion model to generate lip-synced facial images. 
Note that the autoencoders, composed of a decoder $\mathcal{D}(\cdot)$ and an encoder $\mathcal{E}(\cdot)$, are initially pre-trained and frozen in the proposed framework during training. 
 }  
  \label{fig:methods}
\end{figure*}

\subsection{Preliminaries}  

\subsubsection{Problem Definition}  The task of audio-driven lip sync is to generate lip-synced videos given a segment of audio, a video, and a style reference video as inputs.  In this task, the lower-half face of the input video is masked, and then the masked region is inpainted with lip-synced content by the proposed method.

\subsubsection{3D Morphable Model (3DMM)}
According to~\cite{blanz1999morphable}, the 3DMM mainly serves as a comprehensive three-dimensional facial model that represents facial shape using a fixed number of points. 
Specifically, the facial shape can be depicted using a facial mesh $S \in \mathbb{R}^{3N}$ with $N$ vertices, and such a facial mesh $S$ can be parameterized via Principal Component Analysis (PCA) as follows,
\begin{equation}
\label{eq:PCA}
S = \overline{S} + \alpha B_{id} + \beta B_{exp}.
\end{equation}
Here, $\overline{S} \in \mathbb{R}^{3N}$ denotes the average facial shape, while $B_{id}$ and $B_{exp}$ represent the PCA bases for identity and expression, respectively. The parameters $\alpha \in \mathbb{R}^{80}$ and $\beta \in \mathbb{R}^{64}$ control facial identity and expression, respectively. Head rotation is expressed using another set of parameters, $\gamma \in \mathbb{R}^{3}$.

Typically, off-the-shelf 3D face reconstruction models enable the extraction of the parameters $\alpha, \beta, \gamma$ from any real-world facial image. 
It is important to note that of the 64 expression parameters, only 13 are significantly related to mouth motions, as illustrated by~\cite{ma2023styletalk,wang2024styletalk++}. Therefore, in this work, we utilize these 13 mouth-related expression parameters to represent lip motion.

\subsubsection{Method Overview}
An overview of our method is illustrated in Figure~\ref{fig:methods}. 
In this study, we present a novel two-stage framework called \textbf{StyleSync}, which introduces an innovative audio-aware style reference scheme for style-preserving lip sync. Specifically, we first propose a module named Reference-guided Lip Motion Prediction, aimed at predicting lip shapes from input audio signals guided by a style reference video. This is achieved through an innovative Transformer-based model that aggregates speaking style information from the style reference video. Then, \textbf{StyleSync} introduces another module called Realistic Face Rendering, which utilizes a conditional latent diffusion model to produce realistic talking face videos.

\subsection{Reference-guided Lip Motion Prediction}
\label{sec:first_stage}

To accomplish the task of predicting lip motions, we initiate the process with reference audio and lip shape encoding. This step aims to acquire reference audio and lip features from the style reference video. 
Subsequently, we employ the cross-attention layers in transformer decoder to aggregate the speaking style information from these reference audio and lip features. The aggregated style is then employed to guide the lip motion prediction.



\subsubsection{Reference Audio and Lip Shape Encoding}
A style reference video of an individual contains the lip shapes produced when that individual pronounces some utterances, reflecting the speaking style.  
Initially, we employ DeepSpeech~\cite{hannun2014deep} to extract the per-frame DeepSpeech features of the style reference video, denoted as $a^{\prime}_1, a^{\prime}_2, \cdots, a^{\prime}_n$, where $n$ represents the length of the style reference video. 
These DeepSpeech features are further processed by a 1D temporal convolutional module and fed into a Transformer encoder, yielding reference audio features $k_1, k_2, \cdots, k_n$.

To obtain the corresponding lip shape representation from the style reference video, we utilize an off-the-shelf 3D face reconstruction model~\cite{deng2019accurate} to extract 3DMM expression parameters from video frames. We then utilize the mouth-related expression parameters to represent the reference lip shapes, denoted as $m^{\prime}_1, m^{\prime}_2, \cdots, m^{\prime}_n$. These mouth-related expression parameters are inputted into another Transformer encoder, generating reference lip features $v_1, v_2, \cdots, v_n$.


\subsubsection{Audio-Aware Style Aggregation in Transformer} 
Once obtaining the speaking style information from the style reference video, represented by reference audio and lip features, we utilize a Transformer decoder~\cite{vaswani2017attention} for generating mouth-related expression parameters corresponding to the input audio. As illustrated in Figure \ref{fig:methods}, the Transformer decoder consists of $N_1$ Transformer blocks, each comprising a stack of self-attention, cross-attention, and feed-forward layers.

To capture temporal coherence, the Transformer decoder takes audio features from a temporal window $t-w, t-w+1, \cdots, t+w$ as input, where $w$ represents the window size and $t$ denotes the current time step. 
Guided by the reference audio and lip features, it predicts the lip motion at the middle time step $t$. 
To achieve this goal, we first encode the DeepSpeech features of the temporal window into audio tokens using a 1D convolutional module, denoted as $a_{t-w}, a_{t-w+1}, \cdots, a_{t+w}$. 
Subsequently, these audio tokens are inputted into the self-attention layer of the transformer decoder to generate hidden features $h_{t-w}, h_{t-w+1}, \cdots, h_{t+w}$.

To achieve style aggregation, previous methods, such as~\cite{ma2023styletalk,wang2024styletalk++,ma2023dreamtalk}, utilize a self-attention pooling layer on reference lip features to aggregate the style information. Unlike these previous approaches, this work proposes a novel audio-aware style aggregation scheme using cross-attention layers.

Specifically, we begin by treating the obtained reference audio features as keys $\mathbf{K}={[k_1, k_2, \cdots, k_n]}^\top \in \mathbb{R}^{n\times d}$, and the corresponding reference lip features as values $\mathbf{V}={[v_1, v_2, \cdots, v_n]}^\top \in \mathbb{R}^{n\times d}$, where $d$ represents the dimension of the key. These keys and values are inputted into the cross-attention mechanism. Simultaneously, the hidden features of temporal window act as queries in the cross-attention mechanism, denoted as $\mathbf{Q}={[ h_{t-w}, h_{t-w+1}, \cdots, h_{t+w} ]}^\top \in \mathbb{R}^{(2w+1)\times d}$. Therefore, the style reference information can be aggregated through the cross-attention layer, formulated as follows,
\begin{equation}
\mathbf{s}=\operatorname{softmax}\left(\frac{\mathbf{Q} \mathbf{K}^\top}
{\sqrt{d}}
\right) \mathbf{V}
\end{equation}
where $\mathbf{s} \in \mathbb{R}^{(2w+1)\times d}$ denotes the aggregated styles for the temporal window at time step $t$. 

It is noteworthy that, when predicting lip motions for different audio inputs, the attention weights $\frac{\mathbf{Q} \mathbf{K}^\top}{\sqrt{d}} \in \mathbb{R}^{(2w+1)\times n}$ vary based on the similarity between the reference audios and the audios of the temporal window. 
The closer the reference audio is to the audio within the temporal window, the more its corresponding reference lip shape contributes to aggregating the style.  
This differs from prior works~\cite{ma2023styletalk,wang2024styletalk++,ma2023dreamtalk,fu2024mimic}, which aggregate a video-level style remaining static for predicting lip motion for arbitrary audio.

The aggregated styles are then added to the hidden features in a residual manner, which are further fed into the feed-forward layer. 
As a result, the Transformer decoder outputs $2w+1$ vectors, among which the middle vector, corresponding to the ($w+1$)-th position, is inputted into an MLP layer to generate mouth-related expression parameters, denoted by $\hat{m}_t$, for the time step $t$.

\subsubsection{Training Objectives}
During the stage of lip motion prediction, the optimization involves L1 reconstruction for both mouth-related expression parameters and lip vertices, as detailed below.

As the proposed StyleSync predicts lip motion for each frame individually, ensuring temporal consistency in Lip Motion Prediction is crucial. To achieve this, we follow the common practices of previous methods\cite{wang2022one,ma2023styletalk,wang2024styletalk++} by first generating the lip motions of successive $L=64$ frames for each video during training. Subsequently, we utilize the following L1 reconstruction loss for mouth-related expression parameters:
\begin{equation}
\mathcal{L}_1=\frac{1}{L} \sum_{i=0}^{L-1} \left| \hat{m}_{t+i}-m_{t+i} \right|_1
\end{equation}
where $m_{t+i}$ represents the ground truth mouth-related expression parameters for the $(t+i)$-th frame.

Additionally, since mouth-related expression parameters may contain some information unrelated to lip shape, we also impose an L1 reconstruction loss on lip vertices. Specifically, the predicted mouth-related expression parameters $\hat{m}_{t+i}$ can be combined with other parameters and then converted into the face mesh vertices 
using Equation (\ref{eq:PCA}). We denote the lip vertices in the facial mesh as $\hat{l}_{t+i}$, and the reconstruction loss on lip vertices can be expressed as:

\begin{equation}
\label{eq:vertices}
\mathcal{L}_2=\frac{1}{L} \sum_{i=0}^{L-1} \left| \hat{l}_{t+i}-l_{t+i} \right|_1
\end{equation}
where $l_{t+i}$ represents the ground truth lip vertices detected by face alignment tools~\cite{bulat2017far}.

Overall, the optimization objective of this stage can be summarized as follows,
\begin{equation}
\mathcal{L}_{rec}=\mathcal{L}_1+\lambda \mathcal{L}_2
\end{equation}
where $\lambda$ is a scalar for loss item balance, empirically set to 300.

\subsection{Realistic Face Rendering via Diffusion} 
\label{sec:second_stage}

To construct the conditional face rendering model tailored for producing lip-synced faces, we first introduce the latent diffusion model (LDM) proposed by~\cite{rombach2022high}. 
LDM originates from the diffusion models~(DDPM)~\cite{ho2020denoising}, which employ a diffusion process to gradually introduce noise into the real data and train a denoising network to reverse the diffusion process. Once trained, the diffusion models can synthesize data by iteratively denoising a normally distributed variable.
In comparison to DDPM, LDM~\cite{rombach2022high} performs diffusion and denoising processes in the latent space of a pre-trained autoencoder composed of an encoder $\mathcal{E}(\cdot)$ and a decoder $\mathcal{D}(\cdot)$, aiming to reduce the computational cost. Specifically, given an input image $x$, the image is first encoded into the latent, i.e. $z_{0}=\mathcal{E}(x)$, and then diffused to $z_{t^{\prime}}$ with time step $t^{\prime}\in \{1,2,\cdots, T^{\prime}\}$. Afterwards, LDM trains a U-Net-based~\cite{ronneberger2015u} denoising network $\epsilon_{\theta}\left(z_{t^{\prime}}, t^{\prime}\right)$ to predict the noise added to the image latent $z_{0}$, with $\theta$ denoting its learnable parameters. Thus,
the training objective of LDM can be formulated as follows,
\begin{equation}
\label{eq:loss_ldm}
\mathcal{L}_{ldm}=\mathbb{E}_{z, \epsilon \sim \mathcal{N}(0,1), t^{\prime}}\left[\left\|\epsilon-\epsilon_{\theta}\left(z_{t^{\prime}}, t^{\prime}\right)\right\|_{2}^{2}\right], 
\end{equation}
where $\epsilon$ is the ground-truth noise added to the image latent $z_{0}$ and $t^{\prime}$ is uniformly sampled from $\{1, \cdots, T^{\prime}\}$.
During inference, LDM progressively denoise a normally distributed variable $z_{T^{\prime}} \sim \mathcal{N}(0,1)$ until it reaches a clean latent $\hat{z}_{0}$, which can be decoded by $\mathcal{D}(\cdot)$ to synthesize image.

In this work, since we focus on synchronizing the lip shape in the lower-half face,  we conduct both the diffusion and denoising processes in the lower half of the encoded latent as illustrated in \Cref{fig:methods}. Simultaneously, to provide additional context, the upper half of the encoded latent alongside the lower-half portion are integrated into the denoising U-Net.
During training, we encode the ground-truth face into $z_0$ which is then diffused to $z_t^{\prime}$ by the noise.
During inference, we mask the lower half of the input face and encode it into $z^m_0$, of which the lower half is diffused to obtain the initial $z_{T^{\prime}}$.

To generate lip-synced talking video, we need to condition the latent diffusion model on the predicted lip motion obtained as detailed in \Cref{sec:first_stage}. Besides, it is also important to involve the reference facial image condition into the diffusion model, aiming to preserve more appearance details of the target subject. 
Next, we will introduce the conditioning mechanisms used to address these challenges. These include integrating mouth-related expression parameters into LDM for lip motion alignment via modulated convolution, and incorporating reference facial image into LDM for high-fidelity generation via spatial cross-attention.


\subsubsection{Aligning Lip Motion via Modulated Convolution}
\label{sec:align_lip}
In the lip motion prediction stage, we have predicted the mouth-related expression parameters representing the lip motion synchronized to the input audio.
To ensure that the synthesized facial image has lip shape aligned with the one reflected by the mouth-related expression parameters, we integrate the lip motion condition into the latent
 diffusion model via modulated convolutional layers, drawing inspiration from StyleGAN2~\cite{karras2020analyzing}.

Specifically, as shown in the \Cref{fig:methods}, the mouth-related expression parameters are first combined with the identity and rotation parameters extracted from the input video frame, aiming to preserve the subject identity and head rotation.
The combined parameters are then encoded to a latent code by a 1D convolutional module. 
The latent code is employed to modulate the convolution weights in the convolutional layers of denoising U-Net. 
Formally, we denote the original convolution weights as $\omega$. The latent code is mapped to a scale set $\Phi=\{\phi_i\}$ by an MLP layer, where $\phi_i$ is the scale coefficient corresponding to the $i$-th input feature map. Then, we compute the weight of modulated convolution $\omega_{ijk}^{\prime}$ as follow,
\begin{equation}
    \omega_{ijk}^{\prime}=\frac{\phi_i\cdot \omega_{ijk}} {\sqrt{\sum_{i, k} {(\phi_i\cdot \omega_{i j k})}^2 +{\epsilon}^{\prime }}}
\end{equation}
where $j$ indicates the convolution weight for the $j$-th output features map, $k$ indicates the spatial footprint of the convolution, and ${\epsilon}^{\prime }$ is a small constant for avoiding numerical issues. 

Finally, we perform the convolution operation with the modulated convolution weight to produce the output features map.


\subsubsection{High-fidelity Generation with Spatial Attention}
To ensure the generated face has high-fidelity appearance of the subject, a reference facial image should be provided as a condition for the denoising network. 
As shown in the \Cref{fig:methods}, the reference image is initially encoded into the latent space as $z_r$. 
To preserve more appearance details from the reference image, we devise an appearance encoder consisting of residual convolution blocks. 
The appearance encoder converts the reference latent $z_r$ into the multi-scale reference features.
Given the efficacy of preserving appearance details from the reference image proven by~\cite{hu2023animateanyone,bhunia2023person}, we then integrate the resulting reference features into the denoising network via spatial cross-attention.
Specifically, the reference features serve as the keys and values for spatial cross-attention layers, while the hidden features after modulated convolutional layer in the denoising network serve as the queries. Finally, the output of the spatial cross-attention layer are added to the hidden features in a residual manner.

\subsubsection{Training and Inference} For optimization,
we train the diffusion rendering model using the objective as defined in ~\Cref{eq:loss_ldm}. 
The latent diffusion model generates the facial image of each frame individually, thus enabling the generation of videos with arbitrary length. As well, for achieving temporal consistency, we adopt the batched sequential training strategy~\cite{wang2022one} to generate five sequential images for each video during training.   
For model inference during testing, the denoising processes for all frames share the same initial random noise. 
Note that we randomly select the reference facial image from the same video during training, but alternatively, choose any facial image reflecting the subject appearance as reference during inference.

\section{EXPERIMENTS}

\subsection{Experimental Setups}

\subsubsection{Datasets} In this study, we conduct experiments on two widely utilized audio-visual datasets: VoxCeleb~\cite{Nagrani17} and HDTF~\cite{zhang2021flow}. The VoxCeleb dataset comprises over 100,000 utterances from 1,251 celebrities, sourced from videos uploaded to YouTube. It showcases a broad diversity in terms of ethnicities, accents, and ages, and is a standard benchmark commonly used in prior work. On the other hand, the HDTF dataset is a high-resolution audio-visual collection consisting of approximately 362 unique videos spanning over 15.8 hours, with resolutions of 720P or 1080P. To facilitate a fair comparison, all comparison methods, along with the proposed StyleSync, are trained on VoxCeleb and tested on both datasets.




\subsubsection{Implementation Details}
The videos are resampled to 25 fps, and the facial images cropped from the video frames are resized to 256$\times$256 resolution. The proposed method is composed of two stages, each trained separately. 
For Reference-guided Lip Motion Prediction, the style reference video is extracted from the talking clip of the same individual at different time steps. Importantly, there is no overlap between the style reference video and the ground-truth video during testing. The length $n$ of the style reference video is set to 256 following previous methods~\cite{ma2023styletalk}. The transformer decoder consists of $N_1=3$ blocks, with the temporal window size $w$ set to 5. The query dimension $d$ is 256, and we utilize the multi-head attention layer with 8 heads. 
In the stage of Realistic Face Rendering, 
the number of time steps for diffusion, denoted as $T^{\prime}$, is 1000. 
The latent space of the pre-trained autoencoder has a spatial resolution of 64$\times$64. 
During inference, we employ the DDIM~\cite{song2020denoising} sampler with 200 steps to accelerate the generation process. The reference facial image is randomly selected from the input video, and all comparison methods use the same reference facial image for a fair evaluation. After obtaining the generated face, we blend it with the background following the previous method\cite{zhong2023identity}, producing final talking face videos.

\subsubsection{Evaluation Metrics}
\label{sec:Evaluation}
In the task of audio-driven lip sync, the evaluation mainly focuses on the accuracy of lip sync and the visual quality of the generated facial video. 
For the assessment of speaking style preservation in lip sync, this study utilizes the widely adopted metric called Lip Landmarks Distance (\textbf{LipLMD})~\cite{chen2018lip}, which quantifies the discrepancies between the generated lip shapes and the ground-truth lip shapes corresponding to the input audio. 
A lower score of LipLMD indicates that the generated lip shapes are closer to the ground-truth lip shapes reflecting the individual's speaking style, signifying a more accurate preservation of the speaking style in the generated lip sync. 
Furthermore, \textbf{SyncScore} \cite{chung2016out} is another commonly used metric for evaluating the synchronization of lip shape with the input audio. 
The SyncScore is computed based on the distance between the audio feature and visual feature in SyncNet~\cite{chung2016out}.
However, since SyncNet does not model the speaking styles of the individuals, SyncScore primarily assesses whether the generated lip shape matches the lip shape conforming to the general speaking style learned from a large-scale audio-visual dataset. 
As previously mentioned, there are
still discrepancies between the lip shapes conforming to the general
speaking styles and those customized to the individual's speaking style. Therefore, it is worth noting that the \textbf{SyncScore is insufficient to assess the speaking style preservation}. 
A better SyncScore only indicates that the distance between audio and visual features in SyncNet is closer, which does not necessarily mean that the generated lip shapes are more accurate and preserve the speaking style better.
Despite reporting results for this metric, we recommend LipLMD as the preferred indicator for evaluating lip sync quality since our study focuses on the task of style-preserving lip sync.

Additionally, in assessing the visual quality of the generated talking face videos, this study employs Peak Signal-to-Noise Ratio (\textbf{PSNR}) and Structured Similarity (\textbf{SSIM}) \cite{wang2004image} for evaluating pixel-level visual quality. Moreover, we assess feature-level visual quality using the Learned Perceptual Image Patch Similarity (\textbf{LPIPS}) \cite{zhang2018unreasonable} and Fréchet Inception Distance (\textbf{FID}) \cite{heusel2017gans}. Compared to pixel-level measurements, the feature-level evaluations are more aligned with the human perception, as observed in \cite{zhang2018unreasonable,zhen2023human}.

\subsubsection{Comparison Methods}
In this work, we compare the proposed method with several representative state-of-the-art (SOTA) subject-generic approaches in audio-driven talking face video generation, including \textbf{StyleTalk} \cite{ma2023styletalk}, \textbf{DiffTalk} \cite{shen2023difftalk},  \textbf{PD-FGC} \cite{wang2023progressive}, \textbf{IP-LAP} \cite{zhong2023identity}, \textbf{PC-AVS} \cite{zhou2021pose}, and \textbf{Wav2Lip}  \cite{prajwal2020lip}. 
\textbf{StyleTalk} \cite{ma2023styletalk} is a representative subject-generic method capable of modeling speaking style. 
However, the code released for StyleTalk is incomplete and unable to generate results, lacking the phoneme label extraction and training code. 
Consequently, we made some minor modifications to its code for comparison purposes. The modified version utilizes DeepSpeech \cite{hannun2014deep} audio features instead of phoneme labels and is trained using the same objectives as ours. Additionally, StyleTalk's image rendering network can only accept a reference facial image to provide subject appearance information, whereas other comparison methods input more conditions, such as the upper-half face at the current time step. 
For a fair comparison, we replaced StyleTalk's image renderer with our diffusion rendering model to generate results. We only compare with StyleTalk in terms of lip sync quality.
The methods including \textbf{DiffTalk}\cite{shen2023difftalk}, \textbf{IP-LAP}~\cite{zhong2023identity}, \textbf{Wav2Lip}\cite{prajwal2020lip}, and \textbf{Ours} all synthesize talking face videos through inpainting the lower half of the face influenced by the input audio and reference facial image. 
Therefore, for quantitative comparison, the lower half of the face in the input video is masked and then these methods reconstruct the masked area. 
The input video before masking serves as the ground truth for metric computation. 
We train \textbf{DiffTalk}~\cite{shen2023difftalk} using the released official code until convergence, but it generates temporally unstable lip motion. 
It relies on additional frame-interpolation techniques to smooth the results, affecting the comparison fairness. Therefore, the frame-interpolation post-processing was not utilized for a fair comparison. 

\subsection{Quantitative Evaluation}

\begin{table*}[t]
\centering
\caption{Quantitative comparisons between the proposed \textbf{StyleSync} and other SOTA methods. 
* indicates that due to the incomplete released code of StyleTalk, we make several modifications during re-implementation. 
\dag ~indicates that the SyncScore only assesses the lip sync quality in terms of the general speaking style and is not suitable as a primary metric. 
$\uparrow$ indicates higher is better, and $\downarrow$ indicates lower is better.
} 
\label{tab:comp_table1}
\resizebox{\textwidth}{!}{%

\begin{tabular}{@{}l|ccccccccccccc@{}}
\toprule
\multicolumn{1}{l|}{\multirow{2}{*}{Method}} & \multicolumn{6}{c}{VoxCeleb} &  & \multicolumn{6}{c}{HDTF} \\ \cmidrule(l){2-7} \cmidrule(l){8-14}
\multicolumn{1}{c|}{} & LipLMD$\downarrow$ & SyncScore\textsuperscript{\dag}$\uparrow$ & PSNR$\uparrow$ & SSIM$\uparrow$ & LPIPS$\downarrow$ & FID$\downarrow$ &  & LipLMD$\downarrow$ & SyncScore\textsuperscript{\dag}$\uparrow$ & PSNR$\uparrow$ & SSIM$\uparrow$ & LPIPS$\downarrow$ & FID$\downarrow$ \\ \midrule
Wav2Lip~\cite{prajwal2020lip} (\textit{MM'20}) & 0.0180 & \textbf{9.49} & 24.34 & 0.82 & 0.168 & 73.2 &  & 0.0178 & \textbf{9.59} & 22.41 & 0.79 & 0.233 & 88.19 \\
PC-AVS~\cite{zhou2021pose}  (\textit{CVPR'21})  & 0.1042 & 8.34 & 22.07 & 0.77 & 0.128 & 50.0 &  & 0.6545 & 8.88 & 19.13 & 0.68 & 0.183 & 55.83 \\
IP-LAP~\cite{zhong2023identity} (\textit{CVPR'23}) & 0.0121 & 3.53 & 24.20 & 0.84 & 0.114 & 40.4 &  & 0.0118 & 3.56 & 22.37 & 0.83 & 0.119 & 34.35 \\
PD-FGC~\cite{wang2023progressive} (\textit{CVPR'23}) & 0.1529 & 6.41 & 20.26 & 0.70 & 0.165 & 54.8 &  & 0.1550 & 6.57 & 17.83 & 0.65 & 0.207 & 65.39 \\
DiffTalk~\cite{shen2023difftalk} (\textit{CVPR'23}) & 1.3870 & 1.49 & 22.43 & 0.74 & 0.119 & 42.0 &  & 0.5554 & 2.27 & 22.04 & 0.74 & 0.118 & 30.74 \\
StyleTalk*~\cite{ma2023styletalk}  (\textit{AAAI'23})  & 0.0119 & 5.37 & - & - & - & - &  & 0.0125 & 5.56 & - & - & - & - \\
\textbf{StyleSync (Ours)} & \textbf{0.0109} & 7.45 & \textbf{26.69} & \textbf{0.87} & \textbf{0.075} & \textbf{33.8} &  & \textbf{0.0108} & 6.02 & \textbf{24.90} & \textbf{0.85} & \textbf{0.089} & \textbf{26.45} \\ \bottomrule
\end{tabular}%

}
\end{table*}

We conduct quantitative comparisons with the state-of-the-art methods in terms of lip sync quality and visual quality. 
Wav2Lip~\cite{prajwal2020lip}, IP-LAP~\cite{zhong2023identity}, DiffTalk~\cite{shen2023difftalk}, and the proposed approach focuses on generating the lower-half face for lip sync while inheriting the upper-half face from the input video. 
To be fair, we compute the visual quality metrics solely based on the lower half of the generated face.  
We report the quantitative comparison results in the \Cref{tab:comp_table1}.

\subsubsection{Lip Sync Quality} 
As detailed by the LipLMD metric, the proposed method outperforms all of the comparison baselines on both VoxCeleb and HDTF datasets. 
This demonstrates its generation of more accurate lip sync that better preserves the speaking styles of the individuals.
Specifically, Wav2Lip, PC-AVS, DiffTalk, IP-LAP, and PD-FGC lag significantly behind our proposed method in LipLMD due to their incapability to model the speaking styles of individuals. 
StyleTalk incorporates the use of style reference videos to model speaking styles. However, it still lags behind our StyleSync because of its inaccuracies in style aggregation. Specifically, StyleSync's LipLMD obtains lower scores than StyleTalk by 8.4\% on VoxCeleb and 13.6\% on HDTF. 
In SyncScore, Wav2Lip and PC-AVS achieve higher scores than the proposed method. This is because Wav2Lip utilizes SyncNet as the lip sync expert to guide the training of its generator, and PC-AVS adopts an audio-visual contrastive learning mechanism similar to SyncNet's.
However, this only indicates that their generated lip shapes are closer to the lip shapes matching the general speaking style rather than the individuals' speaking styles, since the SyncNet for computing SyncScore overlooks modeling the speaking styles of individuals.

\subsubsection{Visual Quality}
As shown in the \Cref{tab:comp_table1}, the proposed StyleSync consistently outperforms other algorithms in terms of visual quality metrics, including PSNR, SSIM, LPIPS, and FID. 
Particularly noteworthy are the improvements observed in the feature-level metrics LPIPS and FID, where our method demonstrates significant enhancements over other approaches. This suggests that our results align more closely with human perception, demonstrating the efficacy of our conditional latent diffusion model in rendering high-fidelity and realistic faces. As well,
DiffTalk devises a latent diffusion framework to generate high-fidelity talking face videos but still lags behind ours. 
This may be because it simply concatenates the misaligned reference image latent as condition while our method integrates the multi-scale reference image features through spacial cross-attention layers. Furthermore,
IP-LAP employs the predicted optical flow to align the reference image and features but still achieves inferior performance than our algorithm in terms of visual quality. This may be due to the training instability and mode collapse of its GAN-based framework.


\subsection{Qualitative Evaluation}

\begin{figure*}[t]  
  \centering 
  \includegraphics[width=0.90\textwidth]{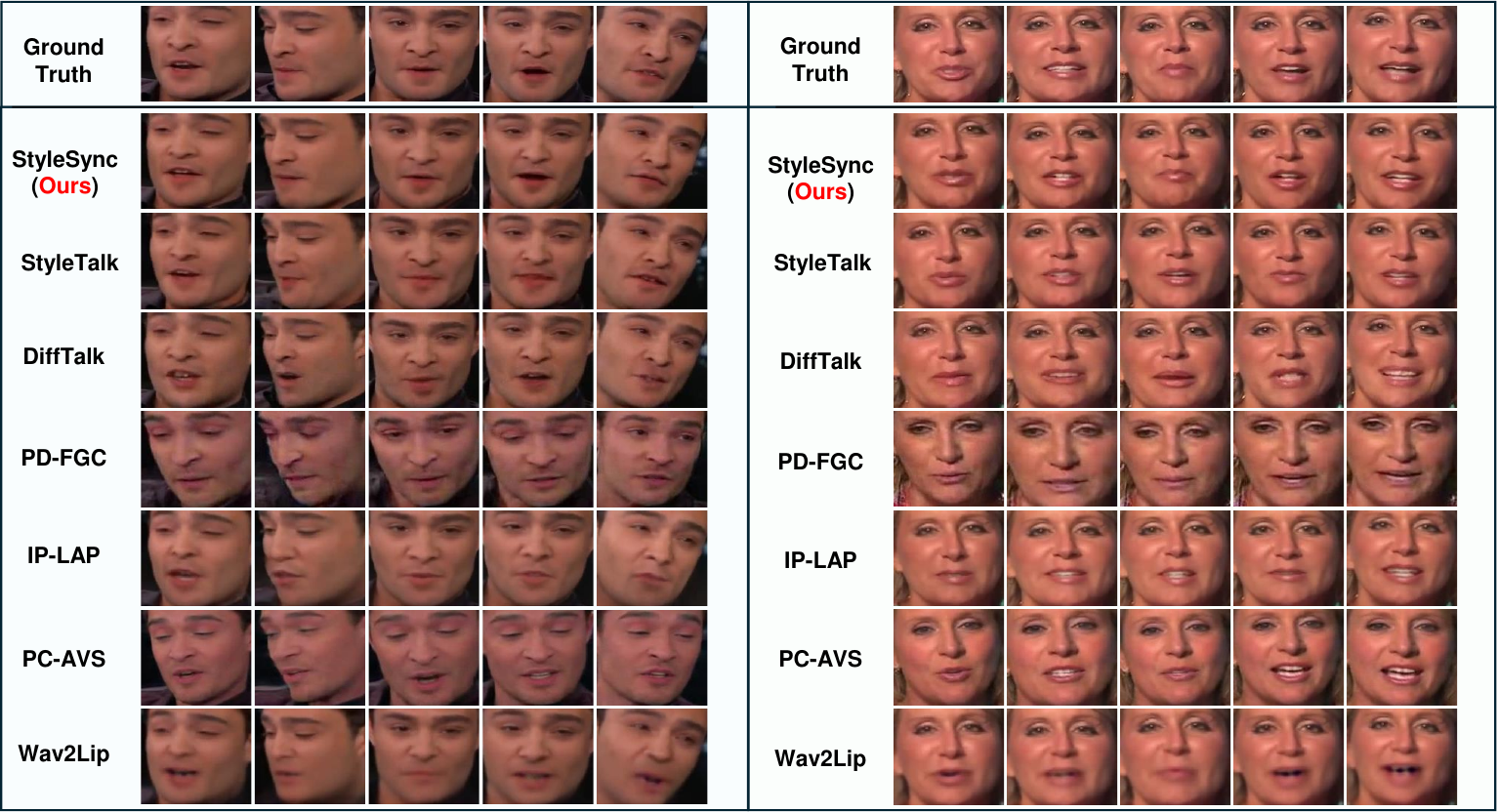} 
  \caption{
  Qualitative comparisons between the proposed \textbf{StyleSync} and other methods, with subjects from the VoxCeleb (Left) and HDTF (Right) datasets. 
  }  
  \label{fig:quality_comp}
\end{figure*}

For a comprehensive qualitative evaluation, we visualize some representative comparison results in~\Cref{fig:quality_comp}. To facilitate comparisons, we present the cropped faces from the generated results of all methods including the ground-truth face. 
Moreover, we conduct a user study from volunteers and provide a supplementary video for further evaluation.

\subsubsection{Lip Sync Quality}
As shown in \Cref{fig:quality_comp}, compared to other algorithms, the proposed method can produce lip shapes that are closer to the ground truth. This indicates our superiority in achieving accurate lip sync conforming to the speaking styles of individuals. StyleTalk sometimes generates inaccurate lip shapes due to its inaccuracy of the style aggregation scheme. Moreover, other approaches, such as 
DiffTalk, PD-FGC, IP-LAP, PC-AVS, and Wav2Lip, produce sub-optimal lip shapes because they overlook modeling the speaking styles of individuals.
 
\subsubsection{Visual Quality} From \Cref{fig:quality_comp}, it can be seen that the proposed method can generate more realistic faces with high-fidelity facial details and fewer artifacts. 
Compared to other methods, our results visually resemble real faces more closely.
This verifies the effectiveness of the proposed conditional latent diffusion model in rendering lip motion into realistic faces.  
During implementation, StyleTalk uses the same rendering model as ours, resulting in visual quality similar to ours. 
Besides, DiffTalk exhibits more artifacts than our approach and loses some facial details of the subject. 
Moreover, PD-FGC and PC-AVS yield results that compromise the subject's identity and display significant flaws. Meanwhile, IP-LAP generates facial details that are blurrier than ours and exhibits artifacts particularly under extreme head poses. Similarly, Wav2Lip produces blurry results with unrealistic mouth details.

\begin{table}[h]
\centering
\caption{User study results. Scores range from 0 to 5, and higher scores indicate superior performance.}
\label{tab:User_Study} 
\resizebox{0.5\columnwidth}{!}{%
\begin{tabular}{@{}rcc@{}}
\toprule
Method & Lip Sync & Visual Quality \\ \midrule
Wav2Lip & 3.15 & 2.60 \\
PC-AVS & 2.58 & 2.33 \\
IP-LAP & 2.19 & 2.35 \\
PD-FGC & 2.74 & 2.32 \\
DiffTalk & 0.90 & 1.49 \\
StyleTalk & 3.93 & - \\ \midrule
\textbf{Ours} & \textbf{4.33} & \textbf{4.60} \\ \bottomrule
\end{tabular}%
}
\end{table}

\subsubsection{User Study} To further validate the effectiveness of our method, we conduct a user study from 12 volunteers following the common practice of previous methods. 
We randomly generate 6 videos of the HDTF~\cite{zhang2021flow} and VoxCeleb~\cite{Nagrani17} datasets for each method.
The volunteers are asked to give their scores (0-5) for each generated video of all comparison methods regarding lip sync quality and visual quality.
We present the mean opinion scores (MOS) of each method in ~\Cref{tab:User_Study}. 
Our method receives better scores from participants across two dimensions than other methods, indicating the superiority of our method.

\subsubsection{Supplementary Video}
We have submitted a supplementary video in the review system.
If there are any issues downloading the video from the system, the same file can also be downloaded via   
\href{https://drive.google.com/file/d/13B0YzAyPrcondcMTJZ9BgavOMVfEz5ig/view?usp=sharing}
{this single-blind backup link.}

\subsection{Style Aggregation Analysis}
\label{sec:style_analyse}
\begin{figure*}[t]
    \centering
    \includegraphics[width=0.9\linewidth]{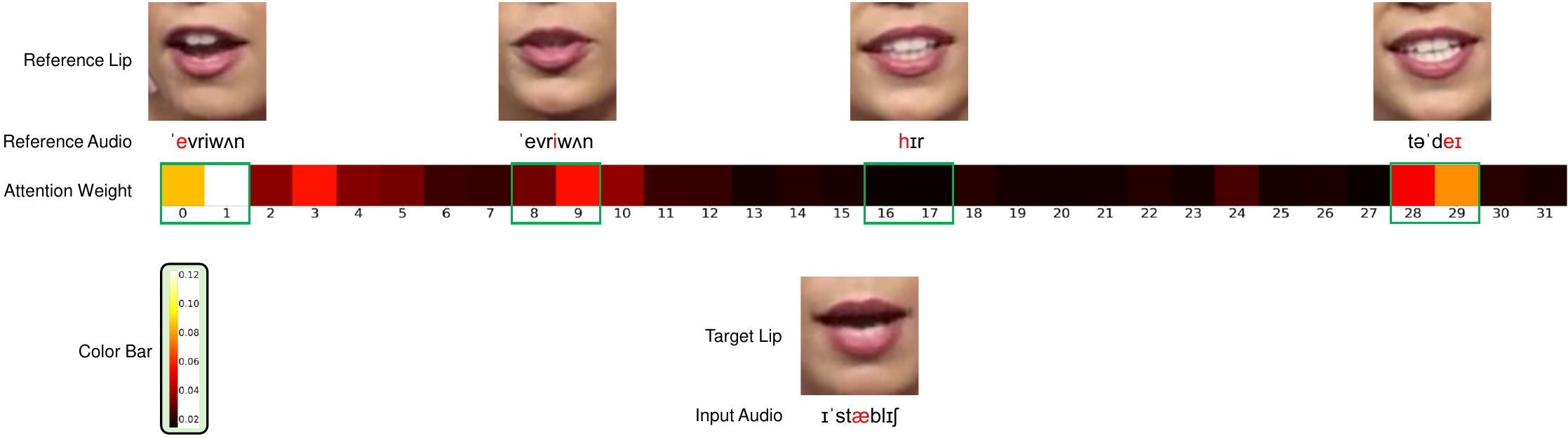} 
    \caption{Visualization of the attention weight in cross-attention layer for style aggregation analysis. 
    The red phonetic symbols represent the content of the audio.
    If a reference lip shape corresponds to the reference audio closer to the input audio, it receives higher attention, contributing more to the style aggregation.}
    \label{fig:analyses} 
\end{figure*}
In our method, we posit that within a style reference video, if a lip shape corresponds to the reference audio that is closer to the input audio, 
it should contribute more significantly to the style aggregation. 
To verify this assumption, we visualize the attention weights of the cross-attention layers in the transformer decoder during the lip motion prediction stage.  There are several cross-attention layers in the transformer decoder. We average the attention weights of all cross-attention layers to obtain a single attention weight, as shown in \Cref{fig:analyses}. To facilitate visualization, we set the length $n$ of the style reference video to 32 frames and denote the frame numbers as $0, 1, 2, \cdots, 31$. 
As illustrated in \Cref{fig:analyses}, the input audio corresponds to the syllable ``æ''. 
The reference audio of the $1$-th frame in the style reference video corresponds to the syllable ``e'', which is closer to the input audio's syllable than the syllables in other frames. 
Therefore, the attention weight for the $1$-th frame is higher than other frames', indicating that the reference lip shape in the $1$-th frame contributes more to the style aggregation.

\begin{table}[t]
\scriptsize
\centering
\caption{Ablation study results of the proposed StyleSync (Ours), with $n$ denoting the length of the style reference video.}
\label{tab:ablation}
\resizebox{0.5\columnwidth}{!}{%

\begin{tabular}{@{}l|c@{}}
\toprule
\multicolumn{1}{l|}{{Variant}} & LipLMD$\downarrow$ \\ \midrule
Ours w/o Reference Audio & 0.01151 \\
Ours w/o Vertices Loss & 0.01145 \\
Ours w/o Reference ($n$=0) & 0.01552 \\
\midrule
Ours ($n$=64) & 0.01125 \\
Ours ($n$=128) & 0.01084 \\ \midrule
\textbf{Ours ($n$=256)} & \textbf{0.01081} \\ \bottomrule
\end{tabular}%

}
\end{table}

\subsection{Ablation Study}

\begin{figure}[]
    \centering
    \includegraphics[width=\columnwidth]{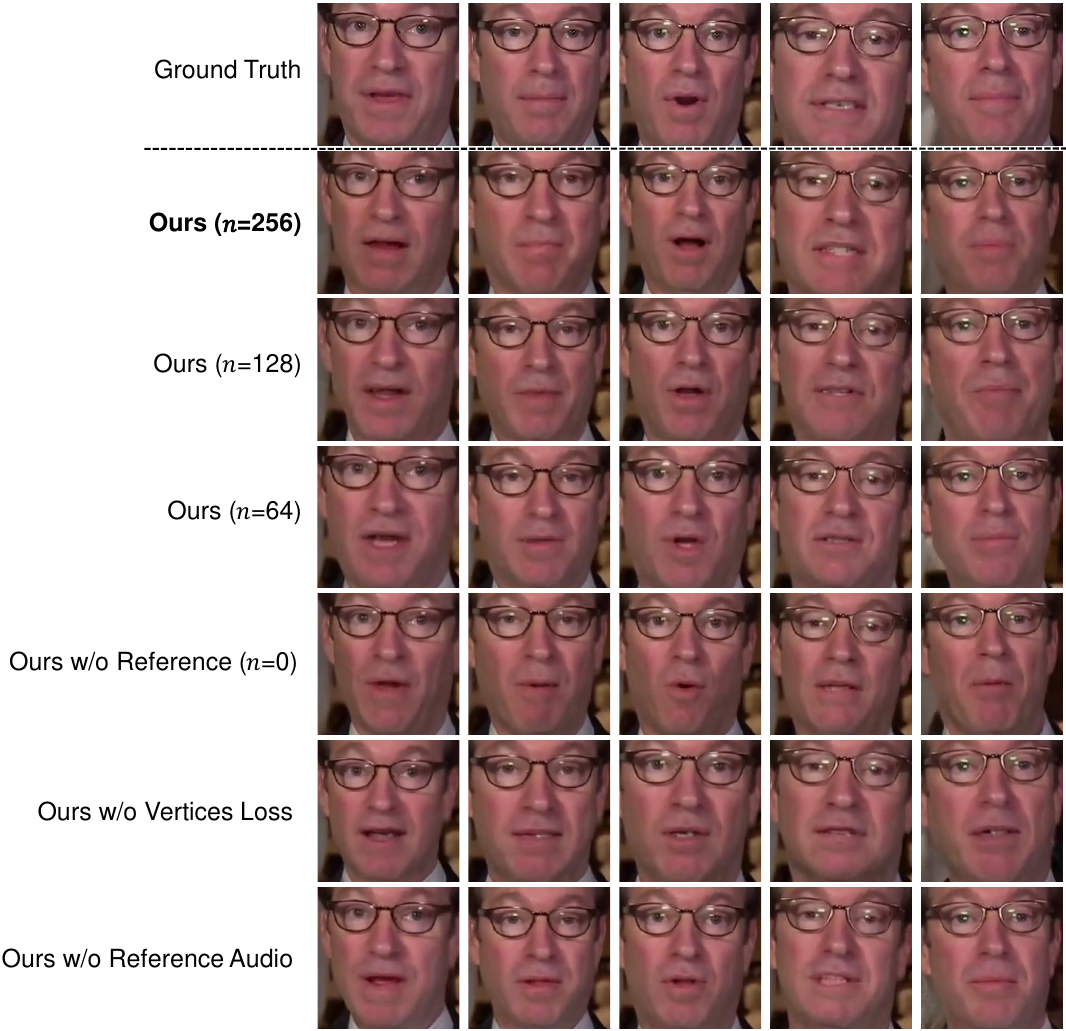}   
    \caption{Qualitative ablation results. Different variants exhibit different lip sync accuracy while retaining the same visual quality. The variation of lip sync quality is more significant in the images of the third column.}
    \label{fig:qual_abla} 
\end{figure}

To understand the advantage of the core components proposed in our StyleSync, we conduct extensive experiments of ablation study on the HDTF dataset. Here, we only adopt the LipLMD metric to evaluate the lip sync quality. This is because the SyncScore primarily assesses whether the generated lip sync matches the general speaking style, as previously mentioned. We report all numerical ablation results in \Cref{tab:ablation} and present the qualitative results in \Cref{fig:qual_abla}.

\subsubsection{Effect of Audio-Aware Style Aggregation} The main contribution of the proposed method is the introduction of the novel audio-aware style reference scheme, which aggregates speaking style information by exploring the relationship between the input audio and the reference audios. Actually, the effectiveness of this has been verified by the fact that the LipLMD score of our method is significantly lower than that of StyleTalk as shown in \Cref{tab:comp_table1}.
Furthermore, to verify the role of the reference audio, we devise a model variant where the reference audios from the style reference video are not utilized for style aggregation like previous methods~\cite{ma2023styletalk,wu2021imitating}. Specifically, the Transformer encoder for encoding the reference audio is removed from the proposed framework, and the reference lip features encoded from the reference lip shapes act as both keys and values for the cross-attention layers in transformer decoder. 
We report the result of this variant in the row "Ours w/o Reference Audio", and 
its LipLMD score increases by 6.5\% compared to our full model's. 
Besides, as shown in \Cref{fig:qual_abla}, without the reference audio, the lip sync quality gets worse. 
This verifies the necessity to explore the relationship between reference and input audio for accurate and style-preserving lip sync. 

\subsubsection{Effect of Vertices Loss}  As the mouth-related 3DMM expression parameters may involve some information unrelated to lip shape, we apply the reconstruction loss over the lip vertices converted from the predicted expression parameters. To verify the effectiveness of this for enhancing lip sync, we devise a variant where the vertices loss in \Cref{eq:vertices} is not utilized for training the lip motion prediction network. The result of this variant is reported in the row "Ours w/o Vertices Loss" of \Cref{tab:ablation}. The LipLMD increases by 5.9\% without using the vertices loss.
As well, from \Cref{fig:qual_abla}, the lip sync accuracy becomes worse without the vertices loss for optimization.

\subsubsection{Effect of Style Reference Video} To evaluate the effectiveness of incorporating the style reference video for modeling the speaking style of individuals, we introduce a variant that predicts lip motion solely from input audio without utilizing the style reference video. In this variant, the reference branch of our lip motion prediction network is removed, and we replace the cross-attention layers in the transformer decoder with self-attention layers. 
The performance of this variant is reported by the row labeled  ``Ours w/o Reference ($n$=0)'' in \Cref{tab:ablation}. It is evident that the LipLMD score significantly increases without modeling the speaking style.


Additionally, to assess the impact of the length $n$ of the style reference video, we varied $n$ during inference and reported results for $n=64$ and $n=128$ in Table \ref{tab:ablation}. Increasing the length of the style reference video leads to improved LipLMD metrics. Besides, as shown in \Cref{fig:qual_abla}, with the length $n$ of the style reference video decreasing, the lip sync quality gradually deteriorates.

\section{CONCLUSION }

In this paper, we have introduced a novel method called \textbf{StyleSync} to tackle the problem of audio-driven lip sync.  
 It can effectively respond to the challenges of traditional methods that can only achieve sub-optimal lip sync conforming to the general speaking styles or struggle to preserve the speaking styles of individuals. 
 Our proposed audio-aware style reference scheme, leveraging the relationships between input audio and reference audio from style reference video, effectively achieves style-preserving lip sync.  
 In detail, the components developed by StyleSync include a novel Transformer-based lip motion prediction model that integrates speaking style information from a style reference video to accurately predict lip motions,  alongside a novel conditional latent diffusion model that translates these predicted motions into realistic, lip-synced videos. 
 Experiments as well as comprehensive ablation analysis have revealed StyleSync’s superiority in audio-driven lip sync.

\bibliography{IEEEabrv,mybib}{}
\bibliographystyle{IEEEtran}


\end{document}